%% file: siamese.tex
\documentclass{article}
\usepackage[preprint]{spconf}
\usepackage{amsmath,graphicx}
\usepackage{fixmath,amssymb}
\usepackage{mathtools}

\usepackage[acronym]{glossaries}
\copyrightnotice{\copyright\ IEEE 2019}
\toappear{{\it Proc.\ ICASSP 2019, May 12-17, 2019, Brighton, United Kingdom}}

\usepackage{float}                  
\usepackage[figuresright]{rotating} 
\usepackage{subcaption}             
\usepackage{graphicx}
\usepackage{eso-pic}                
\usepackage{overpic}                
\usepackage{tikz,pgfplots,pgfkeys}  
\usepackage{psfrag}                 
\usepackage{tabulary}
\usepackage{xparse}                 
\usepackage{todonotes, wasysym} 	  

\usepackage[nodisplayskipstretch]{setspace}

\title{Similarity Learning for Authorship Verification in Social Media}
\name{Benedikt Boenninghoff$~^1$, Robert M. Nickel$^2$, Steffen Zeiler$^1$, Dorothea Kolossa$^1$}
\address{$^1$Faculty of Electrical Engineering and Information Technology, Ruhr University Bochum, Germany\\
  $^2$Department of Electrical and Computer Engineering, Bucknell University, Lewisburg, PA, USA
  }

\newcommand{\norm}[1]{\left\| #1 \right\|_2}

\renewcommand{\vec}[1]{\ensuremath{\boldsymbol{\lowercase{#1}}}}
\renewcommand{\vec}[1]{\ensuremath{\boldsymbol{\lowercase{#1}}}}
\newcommand{\mat}[1]{\ensuremath{\boldsymbol{\uppercase{#1}}}}
\newcommand{\real}{\mathbb{R}}

\renewcommand{\author}{\mathcal{A}}

\newcommand{\loss}{\ensuremath{\mathcal{L}}}

\renewcommand{\^}[1]{\ensuremath{^{(\text{#1})}}}

\begin{document}
\ninept

\newacronym{CNNs}{CNNs}{convolutional neural networks}
\newacronym{RNN}{RNN}{recurrent neural network}
\newacronym{LSTM}{LSTM}{long-short term memory}

\maketitle

\input{abstract/abstract.tex}
\input{introduction/introduction.tex}
\input{theory/theory.tex} 
\input{simulation/simulation.tex}
\input{conclusion/conclusion.tex}

\clearpage
\footnotesize
\vfill\pagebreak
\bibliographystyle{IEEEbib}
\bibliography{ref/refs}

\end{document}

%% file: abstract/abstract.tex
\begin{abstract}
Authorship verification tries to answer the question if two documents with unknown authors were written by the same author or not. A range of successful technical approaches has been proposed for this task, many of which are based on traditional linguistic features such as n-grams. These algorithms achieve good results for certain types of written documents like books and novels. Forensic authorship verification for social media, however, is a much more challenging task since messages tend to be relatively short, with a large variety of different genres and topics. At this point, traditional methods based on features like n-grams have had limited success. In this work, we propose a new neural network topology for similarity learning that significantly 
improves the performance on the author verification task with such challenging data sets.
\end{abstract}


%% file: introduction/introduction.tex
\vspace{-0.1in}
\section{Introduction}
\label{sec:intro}
\vspace{-0.05in}
Social media platforms and text messaging have become a pervasive way of communication in the modern world. A hallmark of such systems is that the true identity of anyone accessing the systems is typically not verified. As a result, users can fall victim to false identity claims. These claims may be made for criminal purposes and/or the distribution of fake news and hate speech for example. An analysis of the authorship of a piece of text can help to reduce the success rate of potentially criminal perpetrators.

Forensic linguistics is the scientific discipline that performs \textit{authorship analysis,} i.e. the
analysis of text documents with respect to authorship, origination, the author's biographical background, and so forth~\cite{Stamatatos09}. The work is usually executed by highly-trained experts.
The term \textit{author profiling} tends to encompass both, the analysis of authorship of documents as well as the author's biographical background. The term \textit{authorship attribution\/} describes a traditional closed-set classification task, where, given a set of candidate authors and documents, the objective is to determine which of the authors has written a set of anonymous or disputed documents~\cite{Juola:2006:AA:1373450.1373451, 7555393}. Lastly, in \textit{authorship verification}, one merely attempts to determine whether two separate documents were written by the same author~\cite{KoppelFundamental, Koppel04}.

Forensic linguistics has developed powerful tools for \textit{authorship analysis,} \textit{author profiling,} \textit{authorship attribution,} and \textit{authorship verification.} The sheer amount of data produced by social media networks and messaging services, however, makes it infeasible to solely rely on trained experts. Engineers have thus begun to develop technical means to process the data by machine or to at least preprocess the data to identify suspicious texts for a later review by experts.
In the last few decades, different computational approaches have been published, especially for authorship attribution. They achieve good results for certain types of written documents such as novels, blog entries or news texts~\cite{Stamatatos09, Juola:2006:AA:1373450.1373451, 7555393, 6638728, Escalante:2011:LHC:2002472.2002510, W17-4907, Ruder2016CharacterlevelAM}. 


An important subtask for many of the proposed algorithms, e.g.~\cite{doi:10.1002/asi.22954, Halvani16, Sapkota15}, is the automatic extraction of linguistic features as a preprocessing step.
In~\cite{Stamatatos09}, for example, stylometric features are categorized into five different groups: lexical, character, syntactic, and semantic features, as well as application-specific features. Some of these, e.g.~character or word n-grams, are easy to extract but more sensitive to a document's content than to its author. Syntactic features, on the other hand, are less sensitive to content but require the use of a robust part-of-speech tagger.
In addition, non-linguistic features such as compression-based models were shown to be successful as well~\cite{838202, Halvani:2017:UCM:3098954.3104050}. The employed compression model, i.e. the \emph{prediction by partial matching\/} approach, requires n-gram counting, however, and is therefore also sensitive to context.

Despite successful approaches in other domains, forensic authorship analysis for {\em social media\/} still remains a great technical challenge, primarily because small sample sizes, i.e. short texts, are quite common and a high variability of genre and topic choice is prevalent~\cite{7555393}. In addition, the writing styles from various social media types, such as email, blog entries, articles, and tweets  differ significantly from those of news texts, which are typically used to train pre-processing tools like a part-of-speech tagger~\cite{Gimpel:2011:PTT:2002736.2002747}.


We are exclusively considering the authorship verification task for social media data in this work, i.e. we investigate similarities in the writing styles for two different social media texts with unknown authors~\cite{KoppelFundamental, Koppel04}. The technical core of our approach is implemented by a hierarchical recurrent Siamese neural network (HRSN). The \gls{RNN} topology allows us to automate the extraction of sensible and context-independent features, even if they may not be linguistically interpretable~\cite{Li15, Yang16}.

The Siamese network topology was originally proposed in~\cite{Bromley:1993:SVU:2987189.2987282} to verify hand-written signatures. Applications in other domains, such as face verification for example, demonstrate the efficiency of the approach~\cite{Chopra05, Hu14, Koch2015SiameseNN, Lu17}. In~\cite{Mueller16, W16-1617} a recurrent Siamese network with \gls{LSTM} cells was proposed to learn semantic similarities between sentences. Recurrent models based on \gls{LSTM} cells have shown a remarkable ability to encode sentences into meaningful, yet compact vector representations~\cite{Sutskever:2014:SSL:2969033.2969173}. In addition, \gls{RNN}s are naturally suited for variable-length sequences, which affords them a structural advantage over convolutional neural network types (CNNs)~\cite{W17-4907, Ruder2016CharacterlevelAM, Kim14, E17-2106}.
Furthermore, hierarchical structures with RNNs at multiple levels were proposed to handle not only sentences but also paragraphs and/or entire documents~\cite{Li15, Yang16}.

The discriminative power of our proposed authorship verification method stems from a fusion of two disparate mechanisms into a single algorithm. Firstly, we are using the Siamese network concept with a specific contrastive loss function to entice maximal separation between same-author/different-author scores during training. Secondly, we are fusing a 2-level hierarchical RNN topology into the Siamese network, which produces fixed-length document representation vectors for variable length texts. The document representation vectors encode those stylistic characteristics of a document that are relevant for authorship verification. Borrowing from the term {\em word embeddings\/} for a collection of vectors that characterize semantic categories for words, we chose to call our representation vectors {\em document embeddings\/} since they encode the stylistic characteristics of entire documents.

%% file: theory/theory.tex
\vspace{-0.15in}
\section{Siamese network Topology}
\label{sec:format}
\vspace{-0.1in}

\begin{figure}[t!]
\centering
\begin{psfrags}
    \psfrag{d}[c][c][.9]{distance measure $d\big(\boldsymbol{x}\^{d}_1, \boldsymbol{x}\^{d}_2\big)$}
    \psfrag{tr}[c][c][.9]{$d\big(\boldsymbol{x}\^{d}_1, \boldsymbol{x}\^{d}_2\big) 
                                \underset{\text{same author}}{\overset{\text{different author}}{\gtrless}} \tau$}
    \psfrag{x1}[c][c][.9]{$\boldsymbol{x}\^{d}_1$}
    \psfrag{x2}[c][c][.9]{$\boldsymbol{x}\^{d}_1$}
    \psfrag{x3}[c][c][.9]{$\boldsymbol{x}\^{w}_1$}
    \psfrag{x4}[c][c][.9]{$\boldsymbol{x}\^{w}_2$}
    \psfrag{s}[c][c][.9]{share}
    \psfrag{t}[c][c][.9]{the same}
    \psfrag{p}[c][c][.9]{parameters}
    \psfrag{n1}[c][c][.9]{Hierarchical}
    \psfrag{n2}[c][c][.9]{RNN topology}
    \psfrag{d1}[c][c][.9]{document 1}
    \psfrag{d2}[c][c][.9]{document 2}
    \psfrag{x1}[c][c][.9]{$\boldsymbol{x}\^{d}_1$}
    \psfrag{x2}[c][c][.9]{$\boldsymbol{x}\^{d}_2$}
    \centerline{\includegraphics[width=6.3cm]{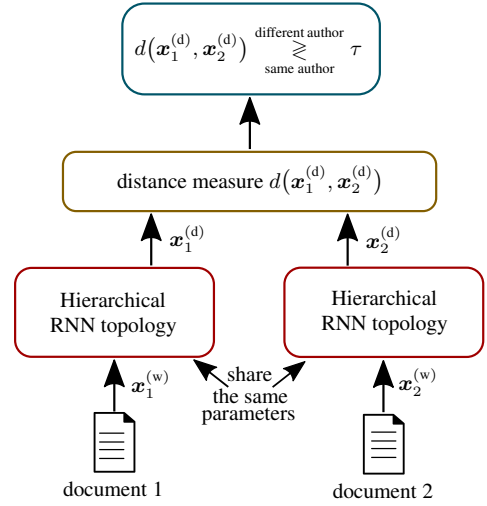}}
\end{psfrags}
\caption{General Siamese network topology.}
\label{fig:siamese}
\vspace{-0.25in}
\end{figure}

Siamese networks consist of two identical neural networks which share the same weights as illustrated in Figure~\ref{fig:siamese}.
Traditional neural networks may learn to classify a given input. Siamese neural networks, however, learn to analyze the similarity of two inputs.
In our case, the input to each RNN is a sequence of word embedding vectors $\boldsymbol{x}\^{w}_i$ for each document $i$ with $i\in \{1,2\}$. Let $\boldsymbol{x}\^{d}_i$ for $i\in \{1,2\}$ be the pair of data points at the output of the \gls{RNN} for each document. We refer to the $\boldsymbol{x}\^{d}_i$ as \emph{document embeddings}.
The output of each sister network is not an easily interpretable object but rather an abstract nonlinear mapping into a high dimensional space. A distance measure, here, the Euclidean distance, is applied to the document embeddings, yielding a similarity measure. A final threshold decision indicates, whether the inputs are similar or not. In the following, we will present our hierarchical   
neural network topology.

\vspace{-0.1in}
\subsection{Hierarchical Recurrent Siamese Network (HRSN)}
\vspace{-0.05in}

\begin{figure*}[!ht]
\centering
\begin{psfrags}
    \psfrag{a}[c][c][.9]{$\text{LSTM}_{\text{w}\rightarrow\text{s}}$}
    \psfrag{b}[c][c][.9]{$\text{LSTM}_{\text{s}\rightarrow\text{d}}$}
    \psfrag{l}[c][c][.9]{$\ldots$}
    \psfrag{h0}[c][c][.9]{$\vec{h}_{0, 1}\^{w}$}
    \psfrag{h1}[c][c][.9]{$\vec{h}_{1, 1}\^{w}$}
    \psfrag{h2}[c][c][.9]{$\vec{h}_{2, 1}\^{w}$}
    \psfrag{h3}[c][c][.9]{$\vec{h}_{T\^{w}\text{-}1, 1}\^{w}$}
    \psfrag{c0}[c][c][.9]{$\vec{c}_{0, 1}\^{w}$}
    \psfrag{c1}[c][c][.9]{$\vec{c}_{1, 1}\^{w}$}
    \psfrag{c2}[c][c][.9]{$\vec{c}_{2, 1}\^{w}$}
    \psfrag{c3}[c][c][.9]{$\vec{c}_{T\^{w}\text{-}1, 1}\^{w}$}
    \psfrag{c4}[c][c][.9]{$\vec{c}_{T\^{w}, 1}\^{w}$}
    \psfrag{h5}[c][c][.9]{$\vec{h}_{0}\^{s}$}
    \psfrag{h6}[c][c][.9]{$\vec{h}_{1}\^{s}$}
    \psfrag{h7}[c][c][.9]{$\vec{h}_{2}\^{s}$}
    \psfrag{h8}[c][c][.9]{$\vec{h}_{T\^{s}\text{-}1}\^{s}$}
    \psfrag{h9}[c][c][.9]{$\vec{h}_{T\^{s}}\^{s} = \vec{x}\^{d}$}
    \psfrag{c5}[c][c][.9]{$\vec{c}_{0}\^{s}$}
    \psfrag{c6}[c][c][.9]{$\vec{c}_{1}\^{s}$}
    \psfrag{c7}[c][c][.9]{$\vec{c}_{2}\^{s}$}
    \psfrag{c8}[c][c][.9]{$\vec{c}_{T\^{s}\text{-}1}\^{s}$}
    \psfrag{c9}[c][c][.9]{$\vec{c}_{T\^{s}}\^{s}$}
    \psfrag{x1}[c][c][.9]{$\vec{x}_{1,1}\^{w}$}
    \psfrag{x2}[c][c][.9]{$\vec{x}_{2,1}\^{w}$}
    \psfrag{x3}[c][c][.9]{$\vec{x}_{T\^{w},1}\^{w}$}
    \psfrag{x4}[c][c][.9]{$\vec{h}_{T\^{w},1}\^{w} = \vec{x}_{1}\^{s}$}
    \psfrag{x5}[c][c][.9]{$\vec{x}_{2}\^{s}$}
    \psfrag{x6}[c][c][.9]{$\vec{x}_{T\^{s}}\^{s}$}
    \centerline{\includegraphics[width=16cm]{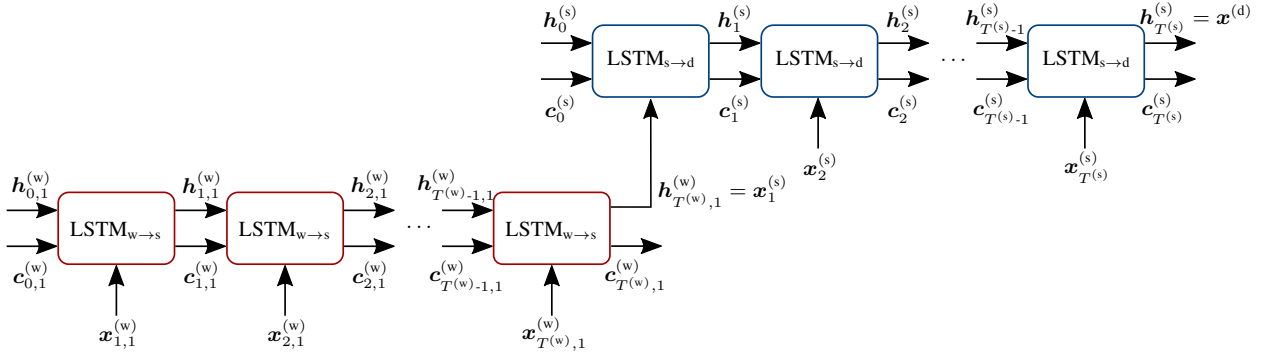}}
\end{psfrags}
\caption{LSTM-based words-to-document encoding scheme.}
\label{fig:RNN}
\vspace{-0.2in}
\end{figure*}

The proposed encoder architecture is illustrated in Figure~\ref{fig:RNN}. 
The main objective is to feed word embeddings into the system in a sentence-by-sentence fashion until each document has been fully encoded. According to~\cite{Li15} we use the 
\gls{LSTM} cell, a type of \gls{RNN} designed to better learn long-term dependencies.

Let $\vec{x}_{t,n}\^{w} \in \real^{D_{\text{w}} \times 1}$ be the $D_{\text{w}}$-dimensional pretrained word embedding of the $t$-th word in the $n$-th sentence. The word-to-sentence encoding scheme can be written as
\vspace{-0.05in}
\begin{align}
    (\vec{h}_{t,n}\^{w}, \vec{c}_{t,n}\^{w}) 
            =\text{LSTM}_{\text{w}\rightarrow\text{s}}\big(\vec{x}_{t,n}\^{w}, \vec{h}_{t-1,n}\^{w}, \vec{c}_{t-1,n}\^{w}\big),
\end{align}
where the hidden state and the memory state are denoted by $\vec{h}_{t,n}\^{w}\in \real^{D_{\text{s}} \times 1}$ and $\vec{c}_{t,n}\^{w}\in \real^{D_{\text{s}} \times 1}$, respectively. $D_{\text{s}}$ defines the dimension of the sentence embeddings. 
The final states are given by $(\vec{h}_{T\^{w},n}\^{w}, \vec{c}_{T\^{w},n}\^{w})$, where the parameter $T\^{w}$ 
denotes the fixed maximum number of words per sentence. 
The first-level update equations are given by
\vspace{-0.05in}
\begin{equation}
\begin{aligned}
    \vec{f}\^{w}_{t,n} &= \sigma(\boldsymbol{W}\^{w}_f~\vec{h}\^{w}_{t-1, n} + \mat{U}\^{w}_f~\vec{x}\^{w}_{t,n} + \vec{b}\^{w}_f), \\
    \vec{i}\^{w}_{t,n} &= \sigma(\mat{W}\^{w}_i ~\vec{h}\^{w}_{t-1, n} + \mat{U}\^{w}_i ~\vec{x}\^{w}_{t,n} + \vec{b}\^{w}_i), \\
    \vec{o}\^{w}_{t,n} &= \sigma(\mat{W}\^{w}_o ~\vec{h}\^{w}_{t-1, n} + \mat{U}\^{w}_o ~\vec{x}\^{w}_{t,n} + \vec{b}\^{w}_o), \\
    \tilde{\vec{c}}\^{w}_{t,n} &= \tanh(\mat{W}\^{w}_c ~\vec{h}\^{w}_{t-1, n} + \mat{U}\^{w}_c ~\vec{x}\^{w}_{t,n} + \vec{b}\^{w}_c), \\
    \vec{c}\^{w}_{t,n} &= \vec{f}\^{w}_{t,n} \odot \vec{c}\^{w}_{t-1, n}+\vec{i}\^{w}_{t,n} \odot \tilde{\vec{c}}\^{w}_{t,n}, \\
    \vec{h}\^{w}_{t,n} &= \vec{o}\^{w}_{t,n} \odot \tanh(\vec{c}\^{w}_{t,n}),
\end{aligned}
\end{equation}
where $\boldsymbol{W}\^{w}_f$, $\boldsymbol{W}\^{w}_i$, $\boldsymbol{W}\^{w}_o$, $\boldsymbol{W}\^{w}_c \in \real^{D_{\text{s}} \times D_{\text{s}}}$,
$\boldsymbol{U}\^{w}_f$, $\boldsymbol{U}\^{w}_i$, $\boldsymbol{U}\^{w}_o$, $\boldsymbol{U}\^{w}_c \in \real^{D_{\text{s}} \times D_{\text{w}}}$
and $\boldsymbol{b}\^{w}_f$, $\boldsymbol{b}\^{w}_i$, $\boldsymbol{b}\^{w}_o$, $\boldsymbol{b}\^{w}_c \in \real^{D_{\text{s}} \times 1}$. 
If the $n$-th sentence is shorter, i.e.~for $\,t_n\^{w} < T\^{w}$, then the hidden state as 
well as the memory state are kept fixed for the remaining iterations, i.e.
\vspace{-0.05in}
\begin{align}
    \vec{h}_{t,n}\^{w} &=\vec{h}_{t_n\^{w},n}\^{w},\\
    \vec{c}_{t,n}\^{w} &=\vec{c}_{t_n\^{w},n}\^{w} \quad \forall t\in \{ t_n\^{w}, t_n\^{w}+1, \ldots , T\^{w} \},
\end{align}
which is readily incorporated into the backpropagation-through-time algorithm during training.

The hidden and memory states are initialized with zero vectors, $\vec{h}_{0, n}\^{w} = \vec{0}_{D_{\text{s}} \times 1}$,
$\vec{c}_{0, n}\^{w} = \vec{0}_{D_{\text{s}} \times 1}$.
The hidden states at index $T\^{w}$ are then used as the compressed sentence embeddings,
\vspace{-0.05in}
\begin{align}
    \vec{x}\^{s}_{n} = \vec{h}\^{w}_{T\^{w},n} \quad \forall n \in \{1,\ldots ,T\^{s}\}.
\end{align}
On the next level, we have the sentence-to-document encoder represented by another LSTM cell,
\vspace{-0.05in}
\begin{align}
    (\vec{h}_{n}\^{s}, \vec{c}_{n}\^{s}) 
            =\text{LSTM}_{\text{s}\rightarrow\text{d}}\big(\vec{x}_{n}\^{s}, \vec{h}_{n-1}\^{s}, \vec{c}_{n-1}\^{s}\big),
\end{align}
where sentence-based hidden and memory states are given by $\vec{h}_{n}\^{s}\in \real^{D_{\text{d}} \times 1}$ and $\vec{c}_{n}\^{s}\in \real^{D_{\text{d}} \times 1}$, respectively. $D_{\text{d}}$ finally defines the dimension of the document embeddings and the final states are given by $(\vec{h}_{T\^{s}}\^{s}, \vec{c}_{T\^{s}}\^{s})$, where the parameter $T\^{s}$ 
denotes the fixed maximum number of sentences per document. 
Analogously to the first level encoder, the second-level update equations are given by
\vspace{-0.05in}
\begin{equation}
\begin{aligned}
    \vec{f}\^{s}_{n} &= \sigma(\boldsymbol{W}\^{s}_f~\vec{h}\^{s}_{n-1} + \mat{U}\^{s}_f~\vec{x}\^{s}_{n} + \vec{b}\^{s}_f), \\
    \vec{i}\^{s}_{n} &= \sigma(\mat{W}\^{s}_i ~\vec{h}\^{s}_{n-1} + \mat{U}\^{s}_i ~\vec{x}\^{s}_{n} + \vec{b}\^{s}_i), \\
    \vec{o}\^{s}_{n} &= \sigma(\mat{W}\^{s}_o ~\vec{h}\^{s}_{n-1} + \mat{U}\^{s}_o ~\vec{x}\^{s}_{n} + \vec{b}\^{s}_o), \\
    \tilde{\vec{c}}\^{s}_{n} &= \tanh(\mat{W}\^{s}_c ~\vec{h}\^{s}_{n-1} + \mat{U}\^{s}_c ~\vec{x}\^{s}_{n} + \vec{b}\^{s}_c), \\
    \vec{c}\^{s}_{n} &= \vec{f}\^{s}_{n} \odot \vec{c}\^{s}_{n-1}+\vec{i}\^{s}_{n} \odot \tilde{\vec{c}}\^{s}_{n}, \\
    \vec{h}\^{s}_{n} &= \vec{o}\^{s}_{n} \odot \tanh(\vec{c}\^{s}_{n}),
\end{aligned}
\end{equation}
where $\boldsymbol{W}\^{s}_f$, $\boldsymbol{W}\^{s}_i$, $\boldsymbol{W}\^{s}_o$, $\boldsymbol{W}\^{s}_c \in \real^{D_{\text{d}} \times D_{\text{d}}}$,
$\boldsymbol{U}\^{s}_f$, $\boldsymbol{U}\^{s}_i$, $\boldsymbol{U}\^{s}_o$, $\boldsymbol{U}\^{s}_c \in \real^{D_{\text{d}} \times D_{\text{s}}}$
and $\boldsymbol{b}\^{s}_f$, $\boldsymbol{b}\^{s}_i$, $\boldsymbol{b}\^{s}_o$, $\boldsymbol{b}\^{s}_c \in \real^{D_{\text{d}} \times 1}$. 
If the document has fewer than the maximum number of sentences, i.e. for $n\^{s} < T\^{s}$, then the hidden and memory state are, again, kept fixed for the remaining iterations, i.e.
\vspace{-0.05in}
\begin{align}
    \vec{h}_{n}\^{s} &=\vec{h}_{n\^{s}}\^{s},\\
    \vec{c}_{n}\^{s} &=\vec{c}_{n\^{s}}\^{s} \quad \forall n\in \{ n\^{s}, n\^{s}+1, \ldots , T\^{s} \}.
\end{align}
The fixed sentence length allows for unrolling the recurrent network and hence, for backpropagation-through-time training.
The hidden and memory states are, again, initialized with zero vectors, $\vec{h}_{0}\^{w} = \vec{0}_{D_{\text{d}}  \times 1}$,
$\vec{c}_{0}\^{s} = \vec{0}_{D_{\text{d}} \times 1}$.
The final hidden states are used as the compressed document embeddings,
\vspace{-0.05in}
\begin{align}
    \label{eq:doc12}
    \vec{x}\^{d} = \vec{h}\^{w}_{T\^{s}}.
\end{align}

\vspace{-0.2in}
\subsection{Distance measure and loss function}
\vspace{-0.05in}
Given a pair of document embeddings, $\vec{x}\^{d}_i$ for $i\in \{1,2\}$ via Equation~(\ref{eq:doc12}), 
we can measure the similarity of both documents by determining the \emph{Eulidean} distance,
\begin{align}
    \label{eq:dist}
    d\big(\vec{x}\^{d}_1, \vec{x}\^{d}_2\big) = \norm{\vec{x}\^{d}_1-\vec{x}\^{d}_2} 
            = \sqrt{\sum_{i=1}^{D\^{d}} \bigg(x\^{d}_{1,i} - x\^{d}_{2,i}\bigg)^2}.
\end{align}
Hence, documents written by the same author result in small values for Equation~(\ref{eq:dist}), while the measure returns large values for documents of different authors.

\begin{figure}[t!]
\centering
\begin{psfrags}
    \psfrag{i}[c][c][1.]{$d($}
    \psfrag{j}[c][c][1.]{$,$}
    \psfrag{k}[c][c][1.]{$)<\tau_1$}
    \psfrag{v}[c][c][1.]{$d($}
    \psfrag{w}[c][c][1.]{$,$}
    \psfrag{q}[c][c][1.]{$)>\tau_2$}
    \psfrag{z}[c][c][1.]{after training}
    \psfrag{n}[c][c][1.]{document written by author $A$}
    \psfrag{m}[c][c][1.]{document written by author $B$}
    \psfrag{a}[c][c][1.]{cross-topic}
    \psfrag{b}[c][c][1.]{same topic}
    \psfrag{x}[c][c][1.]{$\tau_2$}
    \psfrag{y}[c][c][1.]{$\tau_1$}
    \centerline{\includegraphics[width=8cm]{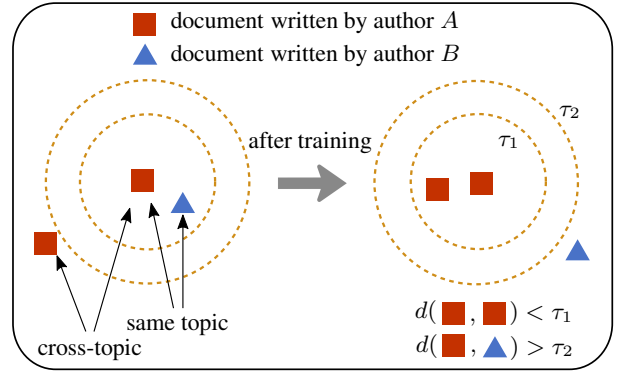}}
\end{psfrags}
\caption{Intuitive illustration of the loss function after~\cite{Hu14}.}
\label{fig:loss}
\vspace{-0.2in}
\end{figure}

The loss function should be chosen carefully. 
In real-world applications, it happens that two documents of the same author may treat diverse topics while two documents written by different authors may consider the same topic.
Such \emph{cross-topic} variations in the dataset leads to missclassifications when using context-sensitive features. 
The left part in Figure~\ref{fig:loss} illustrates this behavior. The distance between 
the cross-topic documents written by author $A$ is larger than the distance between the same-topic documents of author $A$ and $B$. 

Thus, it is desirable to capture context-independent discriminative information through the obtained document embeddings. We can accomplish this by judiciously choosing our loss function during the training phase of the Siamese network.
%
%
For cross-topic/same-author instances, the output of Equation~(\ref{eq:dist}) shall be smaller than a predefined positive threshold $\tau_1$, i.e. $d\big(\vec{x}\^{d}_1, \vec{x}\^{d}_2\big) < \tau_1$. 
On the other hand, for same-topic/different-author instances the output of Equation~(\ref{eq:dist})
shall be larger than a second predefined threshold $\tau_2$, i.e. $d\big(\vec{x}\^{d}_1, \vec{x}\^{d}_2\big) > \tau_2$ with $\tau_1 < \tau_2$. 
Both constraints can be incorporated into a modified version of the \emph{contrastive} loss function~\cite{Lu17},
\vspace{-0.05in}
\begin{equation}
\begin{aligned}
    \loss\big(\vec{x}\^{d}_1, \vec{x}\^{d}_2\big) 
            &= \frac{l}{2} \cdot \max \left\{ d\big(\vec{x}\^{d}_1, \vec{x}\^{d}_2\big) - \tau_1, 0 \right\}^2 
          \\ &~~~+ \frac{1-l}{2} \cdot \max \left\{ \tau_2 - d\big(\vec{x}\^{d}_1, \vec{x}\^{d}_2\big) , 0\right\}^2,
\end{aligned}
\end{equation}
which is applied to all training samples.
The labels are defined as $l \in \{0, 1\}$, where $l=1$ indicates that both texts are written by the same author and $l=0$ means both documents are written by different authors.
After training, the decision threshold may be chosen as 
\vspace{-0.05in}
\begin{align}
    d\big(\boldsymbol{x}\^{d}_1, \boldsymbol{x}\^{d}_2\big) 
                                \underset{\text{same author}}{\overset{\text{different authors}}{\gtrless}} \tau = \frac{\tau_1 + \tau_2}{2}.
\end{align}

%% file: simulation/simulation.tex
\vspace{-0.2in}
\section{Experimental Results}
\label{sec:pagestyle}
\vspace{-0.05in}
In this section, we present experimental results for our proposed authorship verification model with real-world social media data.

\vspace{-0.1in}
\subsection{Dataset}
\vspace{-0.05in}
Currently, there are, unfortunately, not many standardized social media corpora publicly available to reproduce and compare authorship verification results. The dataset employed in this work is based on the one employed by Halvani {\em et al.} in~\cite{Halvani16}, which also incorporates data provided by the PAN evaluation lab~\cite{PAN13, PAN14, PAN15}. It is comprised of a collection of publicly available texts, including samples from different genres like novels, essays, tweets, news articles, social news, product reviews, blog entries, forum posts and emails. The data is specifically designed for authorship verification tasks. It is considered to be particularly challenging, because of its comparatively small size for training and because it contains not only \emph{cross-topic} variations but also handles \emph{cross-genre} conditions, since different text genres are available.


\noindent $\,\,\,\,\,\,$ The accessable dataset (see~\cite{Halvani16, PAN13, PAN14, PAN15}) contains about $9300$ instances of the form $\big(\mathcal{D}_{\text{known}}, d_{\text{unknown}},l\big)$, where $\mathcal{D}_{\text{known}}$ is a set of documents from a known author, $d_{\text{unknown}}$ defines a document in question and $l\in \{0, 1\}$ denotes the class label. The average number of tokens per sentence $16.8$ with a standard deviation of $\pm 13.2$ and the average number of sentences per document is $63.6 \pm 30.2$. A token can refer to a word, abbreviation or punctuation symbol. As mentioned in~\cite{Halvani16}, the number of known documents in each instance varies from one to ten. In~\cite{Halvani16} all documents in $\mathcal{D}_{\text{known}}$ are concatenated into one single document, $d_{\text{known}}$. For the purpose of comparability we decided to do the same. After preprocessing, the average number of sentences per document is finally given by $160.4 \pm 126.3$.


\vspace{-0.1in}
\subsection{Baseline Method}
\vspace{-0.05in}
As our baseline reference, we implemented the authorship verification method published by Halvani {\em et al.}~\cite{Halvani16} in 2016. In~\cite{Halvani16} the authors reported that their approach outperforms two other state-of-the-art methods over the dataset described in Section 3.1.

The baseline method works as follows: a set of character-level features is constructed for each document according to~\cite{Sapkota15}. Features are grouped into an odd number of different categories. A similarity function (see~\cite{Halvani16} for details) is applied to each feature pair from the known/unknown documents. A threshold for each category is used to accept or reject the hypothesis that the document in question belongs to the known author. Finally, a majority vote is applied with respect to the decisions of all feature categories. A training set is needed to find the optimum model parameters, e.g.~the n-gram sizes as well as the equal-error-rate thresholds for all categories.



\vspace{-0.1in}
\subsection{Training Details}
\vspace{-0.05in}
Both algorithms, i.e. our proposed method and the baseline method, were implemented in Python. 
We utilized the library \verb|textacy| for preprocessing to reduce the amount of noise in the data. 
For instance, we removed URLs, email addresses, and phone numbers and replaced them with universal tokens. Specific URLs, email addresses, and phone numbers are typically not part of an author's writing style. 
After text preprocessing we used \verb|spaCy| for sentence boundary detection and tokenization. The training of the neural networks was accomplished with \verb|Tensorflow|. The pretrained GloVe~\cite{Pennington14} word embeddings were taken from \verb|spaCy|. Here, only a small number of $\approx 2\%$ of the segmented tokens were out of vocabulary.
The following training details are mostly inspired by~\cite{Li15, Sutskever14, NIPS2016_6241}:
\begin{itemize}
     \setlength\itemsep{0.02em}
    \item All \gls{LSTM} parameters were initialized from a uniform distribution between $-0.05$ and $0.05$.
    \item The size of the pretrained word embedding was given by $D_{\text{w}}=300$. 
          We then halved the dimensions for each next layer, i.e. $D_{\text{s}}=150$ and $D_{\text{d}}=75$.
    \item The maximum lengths were set to $T\^{w}=33$ and $T\^{s}=123$ to cover $>90$\% of the tokens of a single document in average.
    \item We noticed that using the Adadelta optimizer with a fixed initial learning rate of $1.0$ yields the best results~\cite{Zeiler12}. 
    \item Variational dropout on both \gls{LSTM} levels was set with a rate of $0.3$~\cite{Srivastava14, NIPS2016_6241}.
    \item Gradients were normalized by scaling them when the norm exceeded a threshold of $5$.
    \item A batch size of $32$ was chosen.
    \item $10$-fold cross-validation was applied to test the proposed model. We constructed a development set from the training set such that
         $80\%$ of the data was assigned to the training set, $10\%$ to the development set, and another $10\%$ to the test set. Since the baseline 
         method does not require a development set, its training is performed on the entire training set (including development set).
    \item Data augmentation: We tried to artificially increase the number of training samples. After each epoch, 
          we randomized the order in which documents $\mathcal{D}_{\text{known}}$ were concatenated into a single document $d_{\text{known}}$.
    \item Semi-supervised pretraining: Instead of randomly initializing the trainable \gls{LSTM} parameters we also tried to pretrain the model. 
    In a first step, a sentence-based Siamese network based on~\cite{Mueller16} was trained. For this purpose we constructed a labeled dataset from the SNLI corpus~\cite{Bowman15}.
    In a second step, a hierarchical neural autoencoder was used for unsupervised pretraining of 
    $\text{LSTM}_{\text{s}\rightarrow\text{d}}$~\cite{Li15}. 
    The fixed parameters of the $\text{LSTM}_{\text{w}\rightarrow\text{s}}$ were then given by a previously trained sentence-based Siamese network. 
    According to~\cite{Dai15}, the loss function during the unsupervised pretraining was the Euclidean distance.
    The unsupervised training was performed using the contents of the original Wikipedia paragraphs from~\cite{Li15}.
\end{itemize}

\vspace{-0.15in}
\subsection{Results}
\vspace{-0.05in}
Table 1 summarizes the average verification accuracies over a $10$-fold cross-validation for the proposed Siamese network and the baseline method. 
For the baseline method, we obtained an accuracy of $70.9\%~\pm 1.7$, which is comparable to the average results reported in~\cite{Halvani16}, where Table $7$ shows $\approx 71.1\%$ accuracy for a fixed test set.

Comparing the results of both methods, it can readily be seen that the proposed approach significantly outperforms the baseline system. We were able to increase the average accuracy from around $71\%$ to \mbox{$83.2\% ~\pm 0.8$.} 
An additional improvement was possible with the proposed data augmentation, described in Section 3.3. The resulting increase in the size of our training 
data set led to a further gain of around $2\%$ in performance.
Interestingly, no improvement was accomplished with the proposed pretraining, also described in Section 3.3. The reason for this is likely found in the substantial differences in the nature of the datasets used for pretraining in comparison to the data used for testing. 
%

\vspace{-0.05in}

\begin{table} [htb!]
\centering{ 
\caption{Average precision, recall, F1-scores and verification accuracies for the test set over a $10$-fold cross-validation.}
\vspace{-0.15in}
\def\arraystretch{1.8}
\small
\setlength\tabcolsep{3pt}
\noindent\begin{tabular}{p{1.65cm}||p{1.37cm}|p{1.37cm}|p{1.37cm}|p{1.37cm}||} 
  &\scalebox{.9}{precision} &\scalebox{.9}{recall} &\scalebox{.9}{F1-score} &\scalebox{.9}{accuracy} \\ \hline\hline
    \scalebox{.9}{baseline} &$68.8 ~\pm 2.5$  &$67.8 ~\pm  1.8$  &$68.2~\pm 1.9$  &$70.9 ~\pm  1.7$ \\ \cline{1-5}
    \begin{minipage}{0.1\textwidth}\setstretch{0.3} 
     \scalebox{.9}{HRSN} \end{minipage} & $81.7~\pm 2.4 $ &$81.9 ~\pm 2.7$  &$81.7 ~\pm1.2$ &$83.2 ~\pm 0.8$\\ \cline{1-5}
    \begin{minipage}{0.1\textwidth} \setstretch{0.6} 
   \scalebox{.9}{HRSN (data} \\ \scalebox{.9}{augmentation)} \end{minipage}  
            & $84.3~\pm 2.1 $ &$83.7 ~\pm 2.1$  &$83.9 ~\pm 1.1$ &$85.3 ~\pm 0.9$  \\ \cline{1-5}
    \begin{minipage}{0.1\textwidth}\setstretch{0.3} 
    \vspace*{+0.02in}
     \scalebox{.9}{HRSN} (data \\ \scalebox{.9}{augmentation} \\ \scalebox{.9}{+ pretraining)} \end{minipage}  
                & $84.4 ~\pm 1.6$ & $81.8 ~\pm 1.7$  & $83.0 ~\pm 0.8$ & $84.7 ~\pm 0.8$
\end{tabular}}
\label{tab:acc}
\end{table}

%% file: conclusion/conclusion.tex
\vspace{-0.3in}
\section{Conclusion}
\label{sec:copyright}
\vspace{-0.1in}

We introduced a hierarchical recurrent Siamese network topology for the task of authorship verification. A modified contrastive loss function was chosen during system training to effectively reduce the cross-topic sensitivities of the employed word embeddings. The proposed overall system was shown to be better adjusted than earlier methods~\cite{Halvani16} to the challenges posed by authorship verification with cross-topic and cross-genre social media texts. Improvements of about $15$ percentage points in accuracy were observed.


Further work is still necessary for the development of an effective pretrainig strategy. Here, a better homogeneity between the pretraining data and the training data may be critical.
%
%

In addition, we may go beyond the proof-of-concept presented in this paper and study in greater detail the effects of the loss function with respect to cross-topic and same-topic instances. For this purpose, we intend to compile a larger dataset based on Amazon reviews~\cite{He2016UpsAD} which is labeled w.r.t.~authorship and review categories. Having more data, we are motivated to extend the existing framework: Inspired by~\cite{yang2016hierarchical} we may integrate a three-level attention mechanism for characters, words and sentences. Including character-based representations similar to the ones proposed in~\cite{Ma16} may be beneficial to take the author-specific use of prefixes and suffixes into account. An advantage of the proposed method is, after all, that it is readily modified to incorporate additional text features separately on the character, word, sentence, and document level.

